\documentclass[journal,transmag]{IEEEtran}
\ifCLASSINFOpdf
\else
\fi

\usepackage{graphicx}
\usepackage{amsmath}
\usepackage{bm}
\usepackage{amsfonts}
\usepackage{amssymb}
\usepackage{multirow}
\usepackage{booktabs}
\usepackage{subcaption}

\begin{document}
%
\title{Hard Pixel Mining for Depth Privileged Semantic Segmentation}



\author{\IEEEauthorblockN{Zhangxuan Gu,
Li Niu\IEEEauthorrefmark{*},
Haohua Zhao, and
Liqing Zhang}
\thanks{Zhangxuan Gu is with MOE Key Lab of Artificial Intelligence, Department of
Computer Science and Engineering, Shanghai Jiao Tong University, Shanghai,
China (email: zhangxgu@126.com).

Li Niu is with MOE Key Lab of Artificial Intelligence, Department of
Computer Science and Engineering, Shanghai Jiao Tong University, Shanghai,
China (email: ustcnewly@sjtu.edu.cn).

Haohua Zhao is with MOE Key Lab of Artificial Intelligence, Department of
Computer Science and Engineering, Shanghai Jiao Tong University, Shanghai,
China (email: haoh.zhao@sjtu.edu.cn).

Liqing Zhang is with MOE Key Lab of Artificial Intelligence, Department of
Computer Science and Engineering, Shanghai Jiao Tong University, Shanghai,
China (email: zhang-lq@cs.sjtu.edu.cn).}}

\maketitle

\begin{abstract}
Semantic segmentation has achieved remarkable progress but remains challenging due to the complex scene, object occlusion, and so on. Some research works have attempted to use extra information such as a depth map to help RGB based semantic segmentation because the depth map could provide complementary geometric cues. However, due to the inaccessibility of depth sensors, depth information is usually unavailable for the test images. In this paper, we leverage only the depth of training images as the privileged information to mine the hard pixels in semantic segmentation, in which depth information is only available for training images but not available for test images. Specifically, we propose a novel Loss Weight Module, which outputs a loss weight map by employing two depth-related measurements of hard pixels: Depth Prediction Error and Depth-aware Segmentation Error. The loss weight map is then applied to segmentation loss, with the goal of learning a more robust model by paying more attention to the hard pixels. Besides, we also explore a curriculum learning strategy based on the loss weight map. Meanwhile, to fully mine the hard pixels on different scales, we apply our loss weight module to multi-scale side outputs. Our hard pixels mining method achieves the state-of-the-art results on two benchmark datasets, and even outperforms the methods which need depth input during testing.
\end{abstract}

\IEEEdisplaynontitleabstractindextext

%
\IEEEpeerreviewmaketitle

\section{Introduction}\label{sec:intro}

\IEEEPARstart{S}{emantic} segmentation is a fundamental problem with the goal to classify each pixel in an image, which has a wide range of real-world applications including autonomous driving, visual scene understanding, and image editing. Recently, abundant RGB based semantic segmentation methods~\cite{mostajabi2015feedforward,dai2015convolutional,farabet2013learning,chen2018deeplab,liu2015semantic,zheng2015conditional,jampani2016learning,ye2017salient,vemulapalli2016gaussian,chandra2016fast,wang2019learning} have been developed and achieved excellent performances in given datasets. However, these methods still exhibit clear limitations due to the hard pixels induced by complicated scenes, poor lighting condition, confusing object appearances, and so on.

\begin{figure}
\centering
\includegraphics[width=\linewidth]{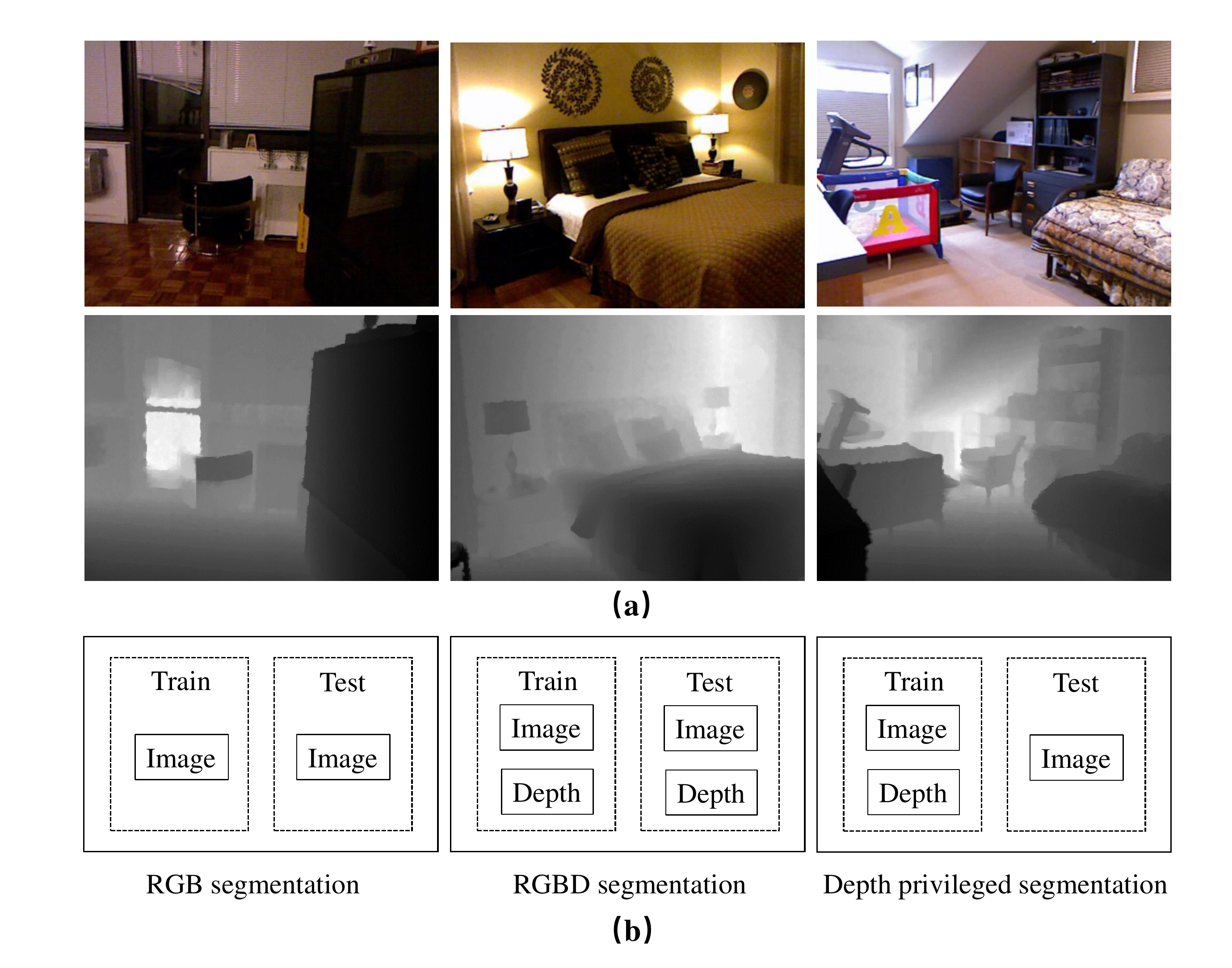}
\caption{(a) Advantages of using depth data over RGB data, where different objects sharing similar RGB appearances have significantly large depth gap. (b) Comparisons between RGB semantic segmentation, RGBD semantic segmentation, and depth privileged semantic segmentation.}
\label{first}
\end{figure}

To address the issues in RGB semantic segmentation, many attempts~\cite{gupta2015indoor,wang2016learning,liu2018collaborative,Hazirbas2016FuseNet,cheng2017locality,Lee2017RDFNet,kang2018depth,li2019joint} have been made to exploit depth information for semantic segmentation, because the depth map can provide complementary 3D information as shown in Fig.~\ref{first}(a), which may be helpful for the segmentation task. For example, \cite{kang2018depth} presents a segmentation network with depth-adaptive multiscale (DaM) convolution layers, while other methods like LS-DeconvNet~\cite{cheng2017locality}, CFN~\cite{Di2017Cascaded}, ACNet~\cite{hu2019acnet}, RDFNet~\cite{Lee2017RDFNet} proposed to fuse RGB features and depth features in different ways. More recently, \cite{li2019joint} utilizes depth to segment and track objects for crowds. These methods use extra depth information for both training and test images, which falls into the scope of multi-view learning. Specifically, each training or test sample consists of two views, \emph{i.e.}, RGB and depth. The experimental results showed that these methods benefit the segmentation tasks from using depth information.

\begin{figure*}
\centering
\includegraphics[width=\linewidth]{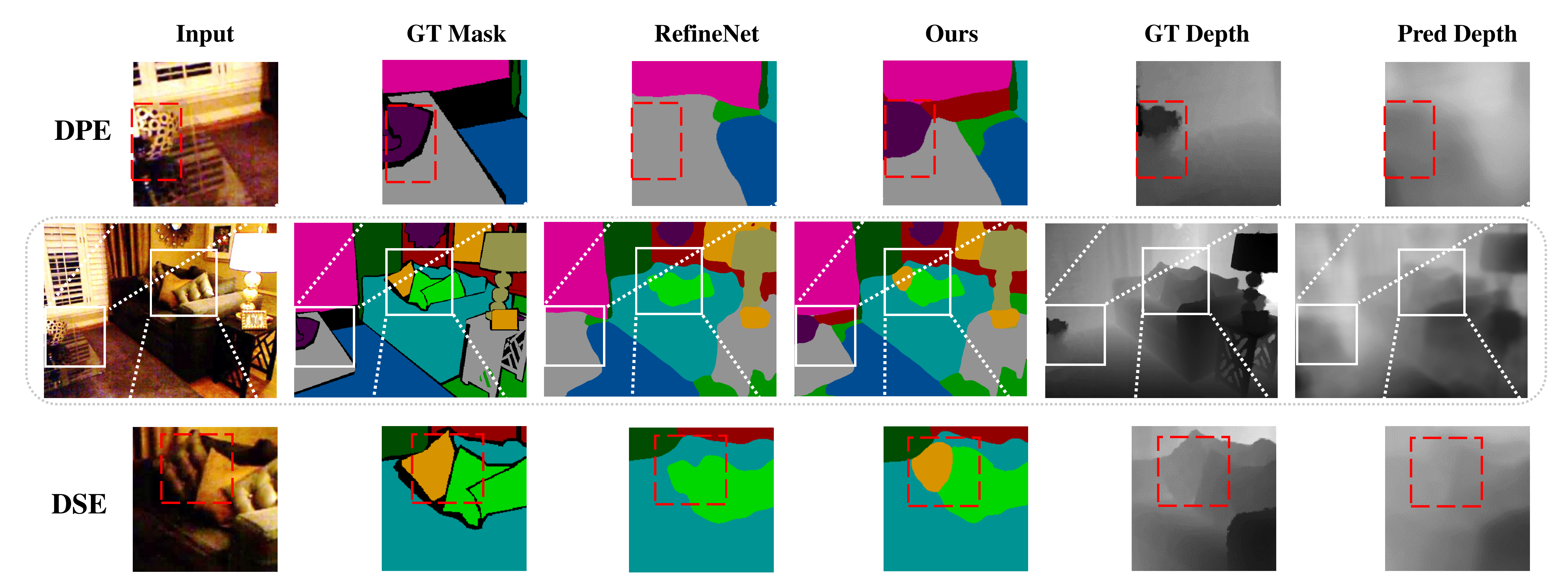}
\caption{Illustration of \textbf{hard pixel measurements} in semantic segmentation. For the entire image and two cropped regions, we show the RGB inputs, the ground-truth segmentation masks, the segmentation results of baseline RefineNet, the segmentation results of our method, the ground-truth depth map, and the predicted depth map of our method from left to right. The first row shows a region with large depth prediction error by comparing ground-truth depth and predicted depth. The chair and the table have a huge depth gap, but the chair is misclassified as a table by baseline RefineNet. The third row shows a region with a similar depth, and the predicted depth is close to the ground-truth depth. The cushion is misclassified as a pillow by baseline RefineNet due to their similar visual appearances. Best viewed in color.}
\label{f1}
\end{figure*}


However, the above methods require depth information of test images during testing, which is often inaccessible in the real-world applications. Therefore,
some other semantic segmentation works attempted to use depth information only in the training stage, in accordance with Learning Using Privileged Information (LUPI)~\cite{Vapnik2009A}, in which privileged information means the information only available for training samples but unavailable for test samples. It has been proved in~\cite{feyereisl2012privileged,fouad2013incorporating,fouad2013ordinal,xu2015distance} that only using privileged information of training samples can still help to learn a more robust model. We refer to the semantic segmentation task using depth as privileged information as depth privileged semantic segmentation. We compare three settings: RGB semantic segmentation, RGBD semantic segmentation and depth privileged semantic segmentation in Fig.~\ref{first}(b). Several multi-task methods like LW-RefineNet~\cite{nekrasov2018lightweight}, TRL~\cite{zhang2018joint}, and PAP~\cite{zhang2019pattern}, which simultaneously handle segmentation task and depth prediction task could be used for depth privileged segmentation. 

One severe drawback of the above depth-privileged segmentation methods is that they treat all pixels in an image equally and exhibit clear limitations due to the hard pixels induced by complicated scenes, confusing object appearances, and \emph{etc}. In fact, one research line of LUPI for image classification is using privileged information to mine hard samples and assigning larger weights on the training losses of these samples~\cite{Sharmanska2014Learning}. Hard sample mining has also been proved effective in previous research works~\cite{Lin2017Focal,Shrivastava2016Training,bulo2017loss} not limited to LUPI. However, no previous works attempt to fuse RGB and depth information to mine hard pixels for the semantic segmentation task.

In this work, we propose a novel network module, \emph{i.e.}, Loss Weight Module (LWM), to mine the hard pixels in semantic segmentation. Measurement of hard pixels is an open and challenging problem. To this end, our proposed LWM employs two hard pixel measurements: Depth Prediction Error (DPE) and Depth-aware Segmentation Error (DSE). The motivation for hard pixel measurements can be explained as follows. By considering segmentation and depth prediction as two joint tasks, we conjecture that segmentation error and depth prediction error are positively correlated. For example, if two neighboring objects from two different categories have a huge depth gap, mispredicted depth may lead to the failure of delineating the depth boundary between them, resulting in the segmentation error of these two objects. In other words, inaccurate depth prediction of these two regions, which leads to the failure of detecting depth boundary, might be highly correlated with the segmentation error. As illustrated in the first row in Fig.~\ref{f1}, a chair is placed next to a glass table with a large depth gap. However, the chair is misclassified as the table, probably because they are visually similar. Therefore, we conjecture that Depth Prediction Error (DPE) could be used as a measurement of hard pixels, which is also quantitatively verified in Section~\ref{correlation}.

Nevertheless, in some circumstances, DPE may be less effective in identifying hard pixels. For example, in a depth-aware local region (a local region with similar depth), if the categories of different subregions are confused with each other due to similar visual appearance, this region becomes a hard region. In this case, accurate depth prediction cannot help separate different subregions, and thus DPE may not be correlated with the difficulty of segmenting this hard region.
As illustrated in the third row in Fig.~\ref{f1}, neighboring pillow and cushion with similar depth form a hard region, in which the cushion is misclassified as a pillow although the depth prediction is correct.
Therefore, we introduce another measurement of hard pixels, that is, Depth-aware Segmentation Error (DSE), to pay more attention to those hard depth-aware local regions. Concretely, we divide the holistic image into depth-aware local regions and compute the segmentation error rate within each region to identify those hard regions.

After identifying the hard pixels, our Loss Weight Module (LWM) outputs a loss weight map by fusing the DPE and DSE maps, in which we investigate several fusion strategies. Then, the loss weight map is used to weight pixel-wise training losses to learn a more robust model. Moreover, our LWM can be easily applied to multi-scale outputs in different network layers for mining the hard pixels on different scales, which can fully exploit multi-scale information. Except for training with easy and hard pixels at the same time, we also explore another training strategy with the proposed LWM, \emph{i.e.}, starting with easy pixels and gradually including hard ones. This training strategy is related to curriculum learning~\cite{Bengio2009Curriculum} and capable of further improving the segmentation performance. Our main contributions are summarized as:
\begin{itemize}
\item This is the first work to mine hard pixels in depth privileged semantic segmentation.
\item We propose a novel Loss Weight Module (LWM), which fuses RGB and depth information to mine hard pixels based on Depth Prediction Error (DPE) and Depth-aware Segmentation Error (DSE).
\item We apply our LWM to multi-scale outputs for mining hard pixels on different scales. Based on LWM, we also explore a training strategy similar to curriculum learning to gradually include hard pixels in the training stage.
\item Extensive experiments demonstrate that our method can achieve the state-of-the-art performances on SUNRGBD and NYUDv2 datasets, and even surpass the methods which require input depth information during testing.
\end{itemize}



\section{Related Works}


In this section, we will conduct a comprehensive review of existing RGB semantic segmentation methods, RGBD semantic segmentation methods, and depth privileged semantic segmentation methods. We will also discuss other related works on learning using privileged information and hard sample mining.

\subsection{RGB semantic segmentation}
Deep learning methods~\cite{liu2015semantic,jampani2016learning,mostajabi2015feedforward,dai2015convolutional,farabet2013learning,zheng2015conditional,vemulapalli2016gaussian,chandra2016fast,wang2016higher} have shown impressive results in RGB semantic segmentation. Most of them are based on the Fully Convolutional Networks (FCN)~\cite{long2015fully}. The extension based on FCN can be grouped into the following two directions: capturing the contextual information at multiple scales and designing more sophisticated decoder. In the first direction, some works combined feature maps generated by different dilated convolutions and pooling operations. For example, PSPNet~\cite{zhao2017pyramid} adopts Spatial Pyramid Pooling which pools the feature maps into different sizes for detecting objects of different scales. Deeplabv3~\cite{chen2017rethinking} proposed an Atrous Spatial Pyramid Pooling by using dilated convolutions to keep the large receptive field.
In the second direction, some works~\cite{Lin2016RefineNet,ronneberger2015u,ding2018context} proposed to construct better decoder modules to fuse mid-level and high-level features. For example, RefineNet~\cite{Lin2016RefineNet} is a multi-path refinement network which fuses features at multiple levels of the encoder and decoder. 
However, all the above methods are based on RGB images, while our approach tends to use a depth map as privileged information to mine hard pixels to boost the performance of semantic segmentation.

\begin{figure*}
\centering
\includegraphics[width=\linewidth]{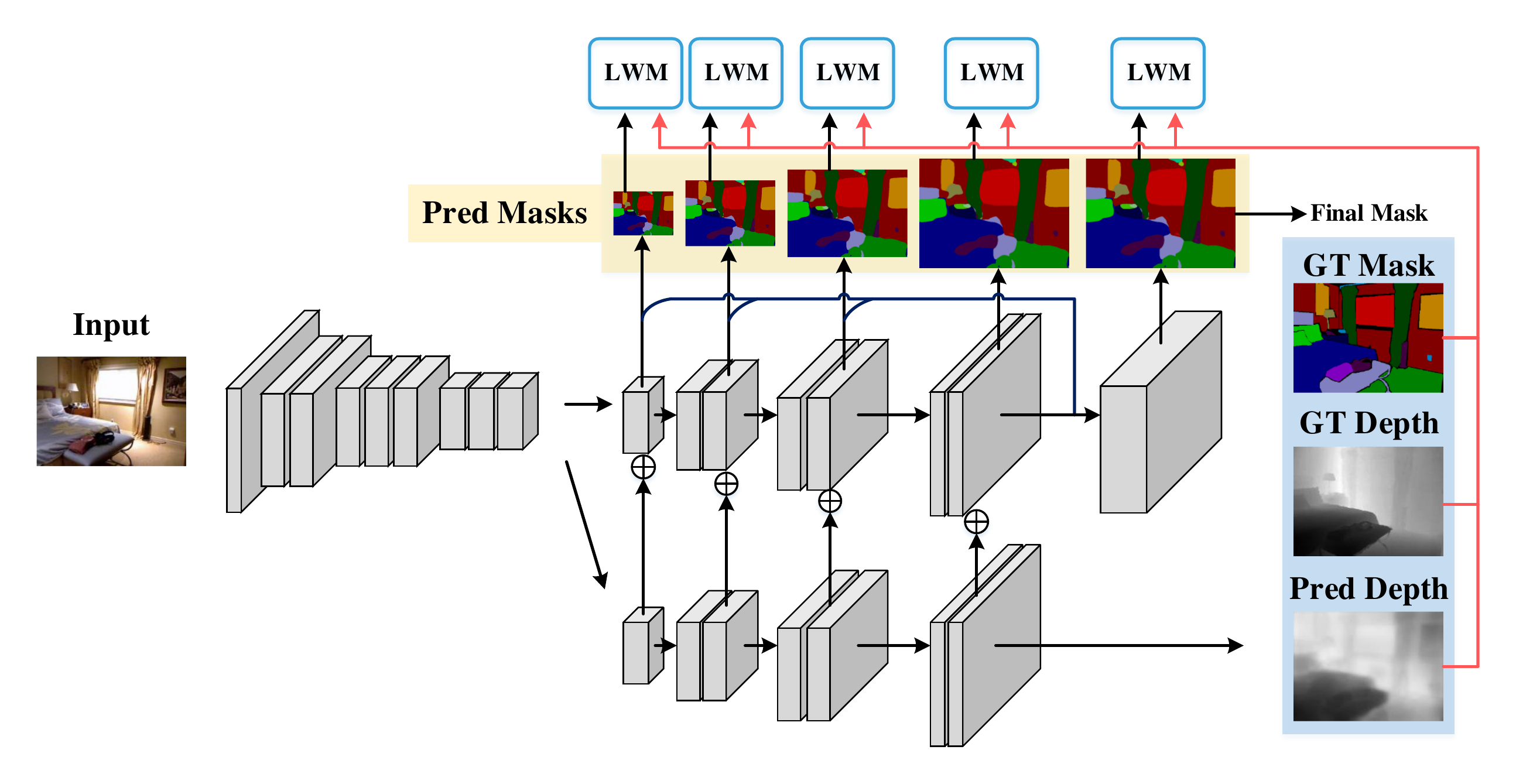}
\caption{An overview of our network architecture. Our method is built upon the multi-task learning framework, in which the segmentation branch and depth prediction branch have one shared encoder branch and two separate decoder branches. We apply our Loss Weight Module (LWM) to the final output as well as four side outputs of the segmentation decoder branch. Our LWM is illustrated in Fig.~\ref{LAM}. Best viewed in color.}
\label{depthbranch}
\end{figure*}

\subsection{RGBD semantic segmentation}

Different from RGB segmentation, recent RGBD segmentation methods exploit depth information under the framework of multi-view or multi-task learning.
By using multi-view learning, many works~\cite{Qi20173D,zhou2018mfdcnn,wang2018depth,cheng2017locality,YuhuaLearning,Di2017Cascaded,hu2019acnet,Lee2017RDFNet,kang2018depth,li2019joint} combine depth information and RGB information in various ways. 
For example, LS-DeconvNet~\cite{cheng2017locality} proposed deconvolution decoder and gated fusion module for RGBD segmentation. Another interesting method ACNet~\cite{hu2019acnet} used Attention Complementary Module to extract and fuse weighted features from RGB and depth branches. More recently, CFN~\cite{Di2017Cascaded} is a neural network with multiple branches of the context-aware receptive field to learn better contextual information, while RDFNet~\cite{Lee2017RDFNet} captures multi-level RGBD features by using the proposed multi-modal feature fusion blocks, which combine residual RGB and depth feature to fully exploit the depth information.

Alternatively, some works~\cite{Jiao2018Look,liu2018collaborative,nekrasov2018lightweight,eigen2015predicting,zhang2018joint,zhang2019pattern} adopt a multi-task learning framework, which predicts depth map and segmentation mask in one unified network.
Although our method is also under the multi-task learning framework, our focus is using depth to mine hard pixels, which contributes more to the segmentation task than merely multi-task learning (see Section~\ref{sec:RDepth}).

\subsection{Learning Using Privileged Information}
The concept of Learning Using Privileged Information (LUPI) was first introduced in~\cite{Vapnik2009A} for classification, in which privileged information stands for the information which is only available in the training stage but not available in the testing stage.
Besides for classification task~\cite{Sharmanska2014Learning,sarafianos2017adaptive,li2014exploiting}, privileged information has also been used for clustering~\cite{feyereisl2012privileged}, verification~\cite{fouad2013ordinal}, hashing~\cite{zhou2016transfer}, random forest~\cite{yang2013privileged}, and \emph{etc}.

Recently, privileged information has also been utilized in deep learning methods~\cite{LambertSS18,yang2017miml,hoffman2016learning}, which aim to distill knowledge from privileged information or use privileged information to control the training process.
More recently, SPIGAN~\cite{lee2018spigan} and DADA~\cite{Vu2019DADA} proposed to use depth as privileged information in the semantic segmentation task. However, their main contribution is exploiting depth information to assist domain adaptation, which adapts the synthetic image domain to the real image domain. So the motivation and solution of their methods are intrinsically different from ours. Some RGBD segmentation methods~\cite{nekrasov2018lightweight,zhang2018joint,zhang2019pattern} under the multi-task learning framework are able to utilize depth as privileged information. 
In contrast with previous depth privileged segmentation methods, this paper is the first work to use depth as privileged information to mine hard pixels for the semantic segmentation task.

\subsection{Hard samples mining}
Abundant previous works have shown that assigning higher weights to the losses of hard samples could help learn a better classifier. Most of them only use one data modality and mine hard samples based on classification score/loss, such as OHEM~\cite{Shrivastava2016Training}, Loss Max-Pooling (LMP)~\cite{bulo2017loss}, and Focal Loss (FL)~\cite{Lin2017Focal}. In the field of LUPI, a few methods like~\cite{Sharmanska2014Learning} use extra modality (\emph{e.g.}, depth) as privileged information to mine hard samples for classification task based on classification loss. However, no previous works have attempted to fuse two modalities to mine hard pixels for the semantic segmentation task.

\section{Methodology}
In this section, we first introduce a multi-task learning framework and then propose our Loss Weight Module (LWM). For ease of implementation, given a training image $\mathbf{I}$, we denote its predicted segmentation mask and ground-truth mask as $\mathbf{S}$ and $\mathbf{S^*}$ respectively. Similarly, we use $\mathbf{D}$ and $\mathbf{D^*}$ to denote its predicted depth map and ground-truth depth map, respectively.

\subsection{Multi-Task Learning}

The multi-task learning framework is based on RefineNet~\cite{Lin2016RefineNet}, which has achieved compelling results in the semantic segmentation task recently. As illustrated in Fig.~\ref{depthbranch}, the encoder consists of four pretrained ResNet~\cite{He2016Deep} blocks, and the decoder consists of four RefineNet blocks. We then use bilinear interpolation to upsample different-scale decoder feature maps to the same size as $640\times480$ and concatenate them in the segmentation decoders. Finally, one $3\times3$ convolutional layer is applied to the concatenated feature map to predict the output segmentation mask. For traditional RGB segmentation, we adopt cross-entropy segmentation loss as follows:
\begin{eqnarray}
L_{s}=-\frac{1}{n}\sum_{i=1}^{n}\sum_{j=1}^{c}y_{ij}log(p_{ij}),
\end{eqnarray}
where $n$ is the number of pixels in image $\mathbf{I}$, $c$ is the number of categories, $y_{ij}$ is the ground-truth binary label to indicate whether the $i$-th pixel belongs to the $j$-th category, and $p_{ij}$ is the predicted probability of the $i$-th pixel for the $j$-th category.

Inspired by~\cite{eigen2015predicting}, we extend another decoder branch to predict depth map, since depth estimation and semantic segmentation are two highly related tasks. To encourage the information to be shared between two decoder branches, we propagate the output feature maps of depth decoder blocks to the output feature maps of segmentation decoder blocks by pixel-wise summation.


For the training loss of depth prediction, inspired by~\cite{wang2015towards}, we compute the mean square error (MSE) of the differences between predicted and ground-truth logarithm of depths. Concretely, given the predicted log-depth map $log\mathbf{D}$ and the ground-truth log-depth map $log\mathbf{D}^*$, we use $logD_i$ and $logD_i^*$ to denote the value indexed by pixel position $i$ in $log\mathbf{D}$ and $log\mathbf{D}^*$ respectively. $D_i$ and $D_i^*$ are in the range of $[1, 256]$, so $log\mathbf{D}$ and $log\mathbf{D}^*$ are in the range of $[0, \log(256)]$. By denoting $R_i = logD_i-logD_i^*$, the depth prediction loss can be written as
\begin{eqnarray}
L_{d}= \frac{1}{n}\sum_{i=1}^n(logD_i-logD_i^*)^2= \frac{1}{n}\sum_{i=1}^nR_i^2.
\end{eqnarray}

%

By combining segmentation and depth prediction loss, the total loss of multi-task learning is $L_s+L_d$ similar to that in~\cite{Nekrasov_2019}. The multi-task learning framework of jointly predicting segmentation mask and depth map is not new for semantic segmentation, but this is not the focus of this paper. Our major contribution is a novel Loss Weight Module (LWM) based on two measurements of hard pixels, which will be introduced in Section~\ref{sec:lwm}. We will also explore applying LWM to multi-scale side outputs in Section~\ref{sec:multi_scale} and a progressive training strategy based on mined hard pixels in Section~\ref{sec:curriculum_learning}.

\subsection{Loss Weight Module (LWM)}\label{sec:lwm}
\begin{figure}
\centering
\includegraphics[width=\linewidth]{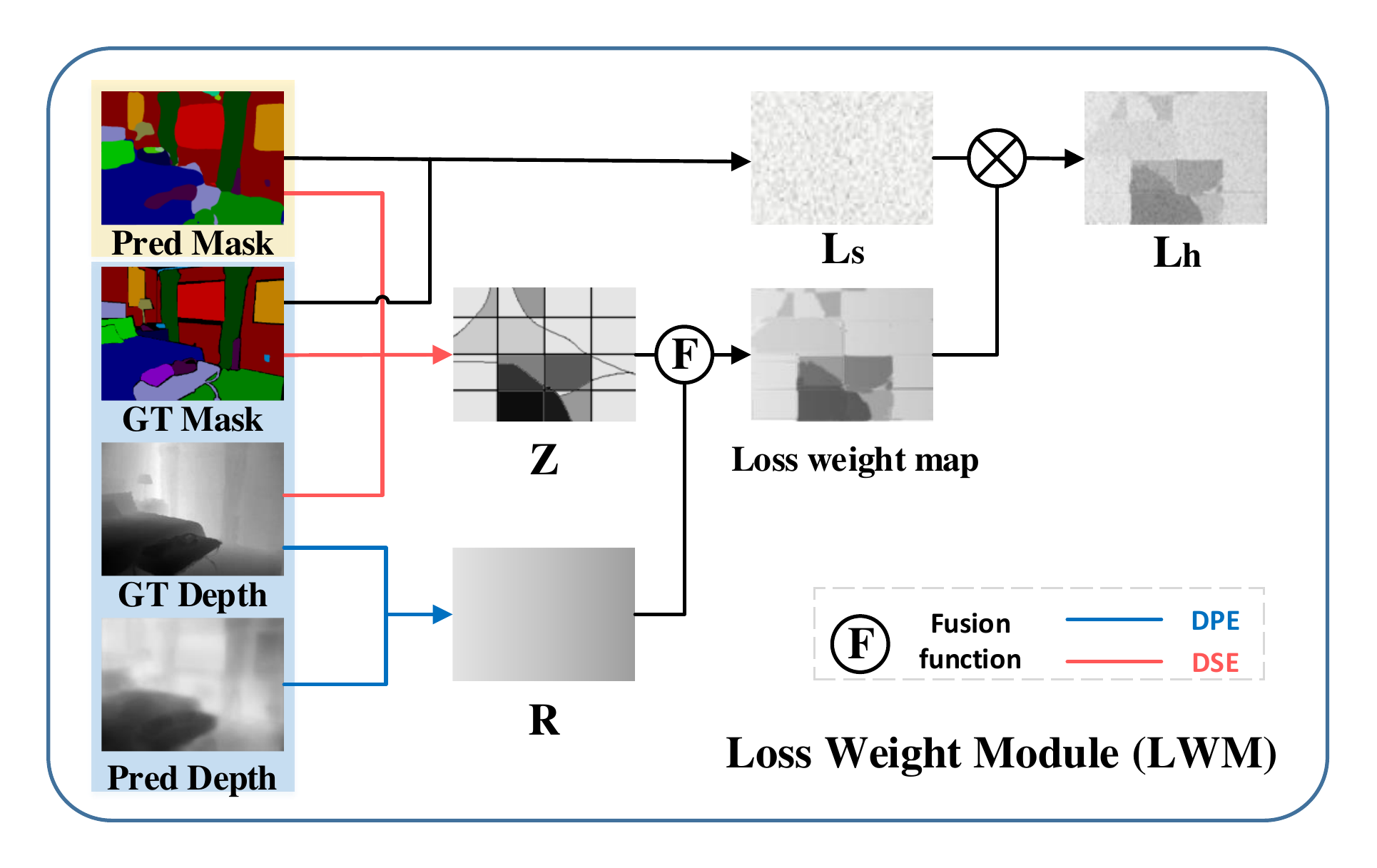}
\caption{Our proposed \textbf{Loss Weight Module (LWM)} takes predicted/ground-truth segmentation masks and depth maps as input, and produces loss weight map. The produced loss weight map is applied to the segmentation loss map by pixel-wise product, leading to the weighted segmentation loss map.}
\label{LAM}
\end{figure}

We tend to identify hard pixels with privileged depth information and assign larger weights to the segmentation losses of hard pixels. The weight map applied to segmentation loss can be deemed as an attention map, which pays more attention to hard pixels to learn a more robust model. Therefore, we refer to our module as Loss Weight Module (LWM), which utilizes two depth-related measurements of hard pixels, \emph{i.e.}, Depth Prediction Error (DPE) and Depth-aware Segmentation Error (DSE), to produce the loss weight map as shown in Fig.~\ref{LAM}. Moreover, this module can be easily integrated into any segmentation network.

\subsubsection{Depth Prediction Error (DPE)} \label{sec:L_R}
As mentioned in Section~\ref{sec:intro} and shown in Fig.~\ref{f1}, accurate detection of the depth boundary between two regions from different categories might be highly correlated with successful segmentation of these two regions. In other words, inaccurate depth prediction of these two regions could cause the failure of detecting depth boundary, which may imply the difficulty of segmenting these two regions.
Thus, we use Depth Prediction Error (DPE) map: $$\mathbf{R} = |log\mathbf{D}-log\mathbf{D^*}|,$$ as a measurement of hard pixels. We normalize DPE within each image to $[0,1]$ by dividing the max value. The correlation between depth prediction error and segmentation error is justified by experiments in Section~\ref{correlation}. Instead of directly using segmentation error as the measurement of hard pixels, the depth prediction error could offer extra guidance after employing additional depth information. Besides, we will introduce another measurement based on segmentation error in the following subsection.

\subsubsection{Depth-aware Segmentation Error (DSE)} \label{sec:DSE}
As discussed in Section~\ref{sec:intro}, when a local region contains different confusing categories within the same ground-truth depth bin, DPE may be less effective in measuring the segmentation difficulty. Thus, we propose another measurement named Depth-aware Segmentation Error (DSE).

\begin{figure}[]
\centering
\includegraphics[width=\linewidth]{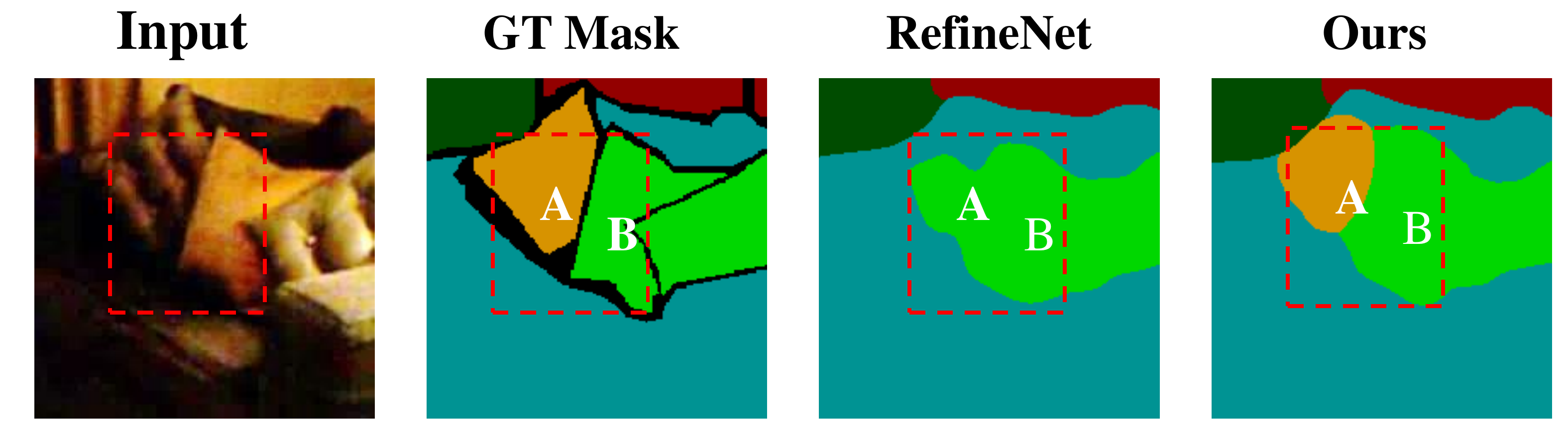}
\caption{The cushion (subregion A) is misclassified as a pillow (subregion B) by baseline RefineNet. We assign higher weights on the whole region AB instead of only the misclassified subregion A.}\label{ABC}
\end{figure}

For a better explanation, we define a local region in the same depth bin as a Depth-aware Local Region (DLR). In some DLRs, the categories of different subregions could be easily confused with each other due to their visual resemblance. We refer to such DLRs with confusing subregions as hard regions. One example is provided in Fig.~\ref{ABC}. The cushion (subregion A) and the pillow (subregion B) have similar depth, and the region AB forms a DLR. The cushion is misclassified as a pillow by RefineNet because the cushion looks very similar to the pillow. Thus, we claim region AB as a hard region. In some other cases, one subregion in a hard region could also be misclassified into an arbitrary wrong category instead of the category of neighboring subregions.
Without loss of generality, we calculate the segmentation error rate in the DLRs as our second measurement DSE. Note that for DSE, we only use the ground-truth depth map $\mathbf{D}^*$ to obtain DLRs without using the predicted depth map $\mathbf{D}$.

To calculate DSE, we need to locate depth-aware local regions (DLRs) first. For ease of implementation, we divide an image into two sets of regions based on two criteria. On one hand, we uniformly divide image $\mathbf{I}$ into $\tilde{w}\times \tilde{h}$ cells, leading to one set of regions $\mathbb{P} = \bigcup_{k=1}^m \{P_k\}$ with $m =\tilde{w}\times \tilde{h}$. On the other hand, we divide the range of ground-truth depth values $\mathbf{D^*}$ (\emph{i.e.}, $[1, 256]$) into $\lceil 256/d \rceil$ depth bins with the size of each bin being $d$ (we empirically set $d=10$ in our experiments, see Section~\ref{sec:hyper}), leading to another set of regions $\mathbb{B} = \bigcup_{t=1}^{\lceil 256/d \rceil} \{B_t\}$, in which each region contains the pixels with depth values $\mathbf{D}^*$ belonging to each depth bin. With two sets of regions $\mathbb{P}$ and $\mathbb{B}$, we calculate the intersection between $P_k$ and $B_t$, yielding a DLR $P_k \cap B_t$. Then, if $P_k \cap B_t \neq \varnothing$, we can simply calculate the segmentation error rate within this DLR as
\begin{eqnarray} \label{eqn:Z}
Z_{P_k,B_t}=(\sum_{i\in P_k \cap B_t}\Delta_i)/|P_k \cap B_t|,
\end{eqnarray}
where $\Delta_i=1$ if pixel $i$ is misclassified and $0$ otherwise. We use $Z_{P_k,B_t}$ to measure the segmentation difficulty of the region $P_k \cap B_t$ and the regions with large $\mathbf{Z}$ are regarded as hard regions. Taking a further step, we use $\mathbf{Z}$ to denote DSE map with the entries in region $P_k \cap B_t$ defined in (\ref{eqn:Z}).
Instead of using segmentation error rate, an alternative choice is calculating weights based on the averaged segmentation loss within each DLR, but that may require a calibration function with additional hyper-parameter like \cite{Lin2017Focal}. So for simplicity, we use segmentation error rates as weights, and the obtained weights are naturally within the range [0,1], which is uniform with our DPE map.

It is worth mentioning that for a hard depth-aware local region (DLR), we assign the same weight to all pixels instead of only misclassified pixels, which can be explained as follows.
Recall that in Fig.~\ref{ABC}, subregion A from category $C_a$ is misclassified as the category $C_b$ . If we only assign higher weights on subregion A, the classifier might be biased towards category $C_a$, which means that a subregion from category $C_b$ bearing visual resemblance with subregion A may be prone to be misclassified as category $C_a$. Therefore, we assume both subregions A and B as the hard region to better distinguish between category $C_a$ and category $C_b$.

One remaining problem is how to determine the number of cells $m$. Two special cases are $m=1$ and $m=w\times h$, in which $w$ and $h$ are the width and height of image $\mathbf{I}$. When $m=1$, the image is divided into depth-aware global regions, in which multiple distant disjoint subregions in the same depth bin could be grouped into one region.
Considering that two distant disjoint subregions in the same depth-aware global region, the segmentation difficulty of one subregion is less affected by the other distant region, so it may be unhelpful to consider them as one hard region. When $m=w\times h$, each cell only contains one pixel. In this case, DSE is equivalent to identifying misclassified pixels instead of hard regions. So depth information is not used, and DSE becomes similar to previous hard pixel mining methods~\cite{Shrivastava2016Training,Lin2017Focal,bulo2017loss} merely based on segmentation error.
In our experiments, we set $m=\bar{w}\times\bar{h}=8\times 8$ by cross validation, which lies between the above two extreme cases (see Section~\ref{sec:hyper}). 

%

\begin{table*}
\centering
\caption{\textbf{Segmentation results} (\%) of different methods on SUNRGBD and NYUDv2. The results marked with * are copied from the corresponding papers and the other results are based on our own implementation. Three types S, D, and P stand for RGB, RGBD, and depth privileged segmentation respectively. ``R'' and ``V'' in the backbone column denote ResNet and VGG respectively. \emph{Note that compared with our method, the methods in the ``D" group use extra depth map of test images during testing and PAP(+SN) uses extra Surface Norm (SN) annotations during training.}}
\resizebox{2\columnwidth}!{
\begin{tabular}{c|c|c|ccc|ccc}
\toprule[1.5pt]
\multirow{2}*{Type}   &\multirow{2}*{Method} & \multirow{2}*{Backbone} &  \multicolumn{3}{c|}{\textbf{SUNRGBD}}    & \multicolumn{3}{c}{\textbf{NYUDv2}}    \\ 
~     & ~  & ~ & \textbf{IoU}  & pixel acc. &  mean acc. & \textbf{IoU}  & pixel acc. &  mean acc. \\ \hline \hline
\multirow{4}*{S}  &RefineNet~\cite{Lin2016RefineNet}& R-50 &   46.73&  78.20   &  64.21 &44.69  & 75.71    &58.53   \\
 ~  &RefineNet~\cite{Lin2016RefineNet}&R-152 & 48.95(45.90*) & 80.81(80.60*)    & 67.45(58.50*) & 46.59(46.50*) &  76.22(73.60*)   &  60.34(58.90*)  \\
 ~  &PSPNet~\cite{zhao2017pyramid}&R-152 &  47.10 &  76.64   &    63.67 &45.91  &  75.31   &  60.18 \\
 ~  &DeepLabv3~\cite{chen2017rethinking}&R-152 & 47.43  & 77.17    & 64.49  &45.52  &  75.58   & 59.70    \\ \hline
\multirow{7}*{P}    &LW-RefineNet~\cite{nekrasov2018lightweight}&R-152 &47.89  &  76.87   &  65.10 & 44.87  & 74.01    &    58.65 \\
~ &RDFNet~\cite{Lee2017RDFNet} (PD)& R-152   & 48.97  &  80.23   & 66.80 &46.78  &  76.21   &   60.12  \\
~ &RDFNet~\cite{Lee2017RDFNet}& R-152    &  48.73 &  78.42   & 66.41 & 46.03 &  75.97  &  59.92   \\
~ &TRL~\cite{zhang2018joint}& R-50   & 49.60*  &  83.60*   & 58.20* &46.40*  &  76.20*   &   56.30*  \\
~ &PAP~\cite{zhang2019pattern} (+SN)& R-50   & 50.50*  &  \textbf{83.80*}   & 58.40* &50.40*  &  76.20*   &   62.50*  \\
~ &\textbf{Ours($L_T$)}&  R-50   & 51.98   & 81.57 & 68.79  & 50.97 & 80.83 &64.01  \\
~ &\textbf{Ours($L_T$)}&  R-152   & \textbf{53.12}   &82.65  & \textbf{70.21}  & \textbf{51.51} &\textbf{81.46}  & \textbf{65.24} \\ \hline
\multirow{6}*{D}   & FCN~\cite{long2015fully}&  V-16     &  31.76   &  61.73     & 49.74 & 34.00*    & 65.40*      &46.10* \\
~ &LS-DeconvNet~\cite{cheng2017locality}&V-16 &  44.72 & 74.38    & 58.00* & 45.90*    & 71.90*      &60.70*      \\
~ &Wang \emph{et al.}~\cite{wang2018depth}& R-152 & 47.23 & 78.31    & 58.63 & 47.96(48.40*)    & 75.48(-*)      &60.73(61.10*)     \\
~ &CFN~\cite{Di2017Cascaded}&R-152  &  48.10* & -    & - & 47.70*    & -      &-      \\
~ &ACNet~\cite{hu2019acnet}&R-50  &  48.10* & -    & - & 48.30*    & -      &-      \\
~ &RDFNet~\cite{Lee2017RDFNet}& R-152   &  50.20(47.70*) & 81.34(81.50*)    & 67.66(60.10*) & 49.01(50.10*) &  76.77(76.00*)   &  61.32 (62.80*)   \\\bottomrule[1.5pt]
\end{tabular}}
\label{stateofart}
\end{table*}

\subsubsection{Fusion of DPE and DSE} \label{sec:fusion_strategy}
Since DPE map $\mathbf{R}$ and DSE map $\mathbf{Z}$ are in the same scale $[0,1]$, we can use a function $F(\cdot)$ to fuse two maps, yielding the integrated loss weight map $\mathbf{M}=F(\mathbf{R},\mathbf{Z})$. When applying the obtained loss weight map $\mathbf{M}$ to segmentation loss map, we can arrive at our weighted segmentation loss:
\begin{eqnarray}
L_{h}=-\frac{1}{n}\sum_{i=1}^{n}\sum_{j=1}^{c}M_iy_{ij}log(p_{ij}), \label{eqn:loss_lh}
\end{eqnarray}
in which $M_i$ is the entry for pixel $i$ in $\mathbf{M}$.
Intuitively, $L_{h}$ aims to place more emphasis on hard pixels to learn a better segmentation network. For the fusion operator $F(\cdot) $, we experiment with three different choices: pixel-wise summation, pixel-wise product, and pixel-wise max, among which pixel-wise summation achieves the best performance (see Section~\ref{sec:hyper}).

\subsection{LWM on Multi-Scale Side Outputs} \label{sec:multi_scale}
It is commonly acknowledged that the output feature maps of different decoder blocks contain the information of different granularities, which make different contributions to the semantic segmentation task. To take full advantage of our LWM, we apply LWM to multiple decoder blocks to mine hard pixels based on the information of different granularities, as shown in Fig.~\ref{depthbranch}.

Specifically, we add a $1\times 1$ convolution layer to each of the four decoder blocks for predicting multi-scale side outputs (\emph{i.e.}, segmentation masks). Then, we apply LWM loss on four side outputs to better mine the hard pixels and provide better supervision. We denote the weighted segmentation losses on multi-scale side outputs as $\{L_{hk}|_{k=1}^4\}$. We totally produce four side outputs (see Fig.~\ref{depthbranch}) and $L_{hk}$ means employing $L_h$ in (\ref{eqn:loss_lh}) on the $k$-th side output.

So far, our total training loss for \textbf{H}ard \textbf{P}ixels \textbf{M}ining can be written as
\begin{eqnarray} \label{eqn:HPM}
L_{HPM} =  L_{s} + L_{d} + \alpha L_{h}+\beta\sum_{k=1}^4L_{hk} ,
\end{eqnarray}
where $\alpha$ and $\beta$ are two trade-off parameters. Note that we only use the losses based on side outputs during training. In the testing stage, we simply use the final predicted segmentation mask for evaluation.

\subsection{Curriculum Learning Based on LWM}\label{sec:curriculum_learning}
Based on the loss function in (\ref{eqn:HPM}), with identified hard pixels, one intuitive training strategy is employing $L_{h}$ and $L_{hk}$ on all pixels. An alternative strategy is that $L_{h}$ (\emph{resp.}, $L_{hk}$) is first employed on easy pixels and then extended to hard pixels, similar to curriculum learning~\cite{Bengio2009Curriculum}.

Curriculum learning has achieved great success in a variety of applications~\cite{zhang2017curriculum,lotter2017multi,li2017multiple,graves2017automated,pentina2015curriculum}, in which the training process starts with easy training samples and gradually includes hard ones. The key problem in curriculum learning is how to define easy and hard training samples, which is solved by our LWM. One popular curriculum learning approach is Self-Paced Learning (SPL)~\cite{Kumar2010Self}, in which the training samples with small training losses are regarded as easy training samples. Specifically, SPL aims to learn the model parameters $\bm{\theta}$ and the binary indicator vector $\mathbf{v} = [v_1,\ldots, v_{N}]^T$ indicating easy training samples, in which $N$ is the total number of training samples. The objective function of SPL can be written as
\begin{eqnarray}\label{eqn:spl_1}
\min_{\bm{\theta}, \mathbf{v}\in[0,1]^{N}} \sum_{i=1}^{N} v_iL(\mathbf{x}_i;\bm{\theta})-\eta\sum_{i=1}^{N} v_i,
\end{eqnarray}
\noindent in which $L(\mathbf{x}_i;\bm{\theta})$ is the training loss of the $i$-th training sample $\mathbf{x}_i$ and $\eta$ is the learning pace. The problem in (\ref{eqn:spl_1}) can be solved by updating $\bm{\theta}$ and $\mathbf{v}$ alternatingly.
When fixing $\bm{\theta}$, we use $L_i$ to store $L(\mathbf{x}_i;\bm{\theta})$ and $v_i$ can be updated as $\delta(L_i<\eta)$, which is equal to $1$ if $L_i<\eta$ and $0$ otherwise. Then, model parameters $\bm{\theta}$ can be updated by solving
\begin{eqnarray} \label{eqn:spl_2}
\min_{\bm{\theta}} \sum_{i=1}^N \delta(L_i<\eta) L(\mathbf{x}_i;\bm{\theta}),
\end{eqnarray}
which only preserves the losses of easy samples with $L_i<\eta$.
Analogous to (\ref{eqn:spl_2}), we extend $L_{h}$ (\emph{resp.}, $L_{hk}$) to $\hat{L}_{h}$ (\emph{resp.}, $\hat{L}_{hk}$) by filtering out the hard pixels. Next, we take $\hat{L}_{h}$ as an example and $\hat{L}_{hk}$ can be obtained in a similar way. In the current epoch $e$, $\hat{L}_{h}$ is formulated as
\begin{eqnarray} \label{eqn:cl_m}
\hat{L}_{h} = -\frac{1}{n}\sum_{i=1}^{n}\delta(M_i<\eta_e) (\sum_{j=1}^{c}M_iy_{ij}log(p_{ij})),
\end{eqnarray}
\noindent where the learning pace $\eta_e$ is defined as $u^1+e\times(u^2-u^1)/E$ with $E$ being the total number of epochs, $u^1$ (\emph{resp.}, $u^2$) is the median (\emph{resp.}, maximum) of $\{M_1,\ldots,M_n\}$, $\delta(M_i<\eta_e)=1$ if $M_i<\eta_e$ and 0 otherwise. Intuitively, we first choose the easy pixels ($M_i<\eta_e$) for calculating the weighted segmentation loss, and then gradually add hard pixels as the number of epochs increases. Thus, the loss function based on curriculum learning can be written as
\begin{eqnarray} \label{eqn:T}
L_{T} =  L_{s} + L_{d} + \alpha \hat{L}_{h}+\beta\sum_{k=1}^4\hat{L}_{hk}.
\end{eqnarray}

The training strategy based on (\ref{eqn:T}) can be explained as follows. At first, only easy pixels are selected for training with $\hat{L}_{h}$ (\emph{resp.}, $\hat{L}_{hk}$). Then, more and more hard pixels will contribute to training with $\hat{L}_{h}$ (\emph{resp.}, $\hat{L}_{hk}$). Note that when setting $\eta_e$ as a sufficiently large constant, $L_{T}$ will reduce to $L_{HPM}$.

\section{Experiments}

In this section, we compare our method with our special cases and state-of-the-art baselines on two benchmark datasets NYUDv2 and SUNRGBD. We also provide comprehensive ablation studies, hyper-parameter analyses, and in-depth qualitative analyses.

\begin{figure}[]
\centering
\includegraphics[width=\linewidth]{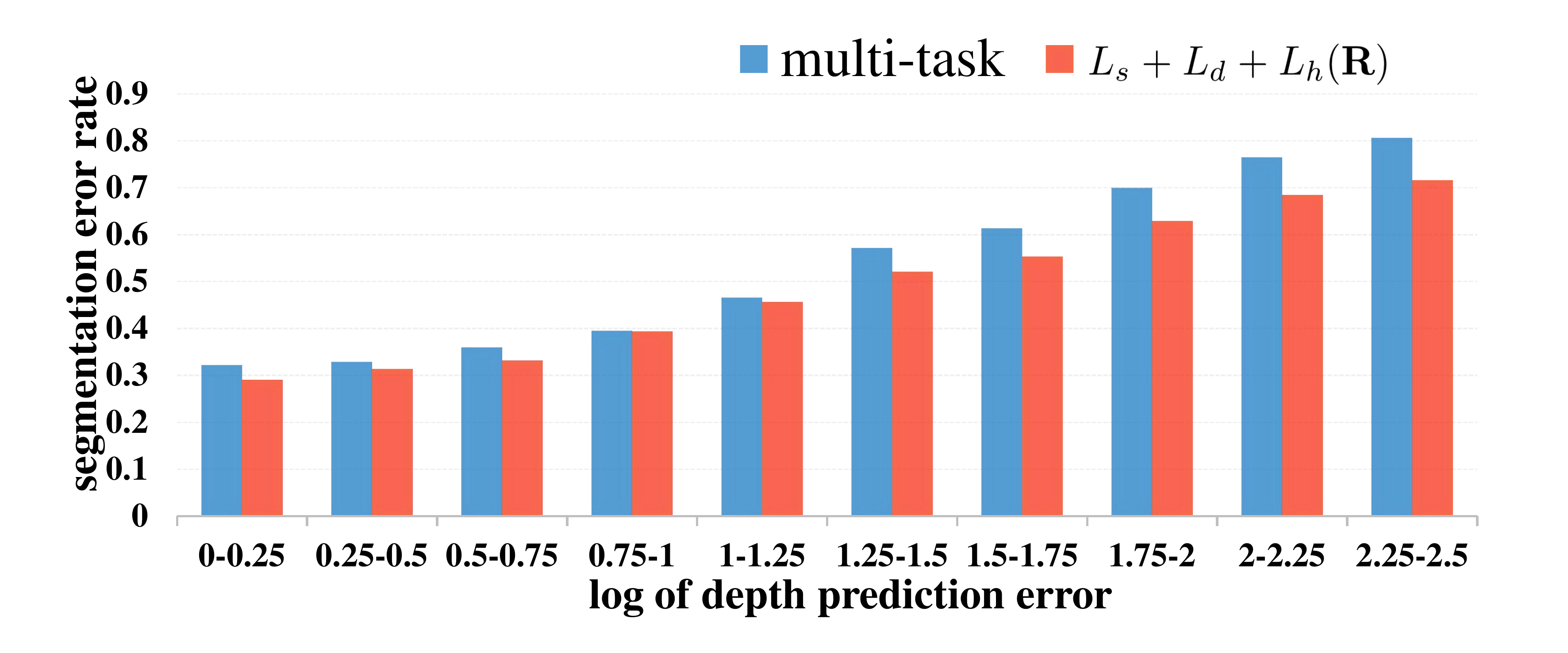}
\caption{\textbf{Correlation} between depth prediction error and segmentation error on NYUDv2. The y-axis is segmentation error rate, while the x-axis is the log-depth-gap bins.}\label{fig:cons}
\end{figure}

\subsection{Datasets and Implementation Details}
We evaluate our method on two benchmark datasets: the popular NYUDv2~\cite{Silberman2012Indoor} dataset and a large-scale SUNRGBD~\cite{Song2015SUN} dataset. The NYUDv2 dataset contains 1449 densely labeled pairs of RGB and depth images. Following the standard train/test split, we use 795 training images and 654 test images. We evaluate our network for 40 categories by using the labels provided by \cite{Gupta2013Perceptual}. The SUNRGBD dataset consists of 10335 pairs of RGB and depth images. We use the standard split of 5285 for training and 5050 for testing with 37 categories.

Our method is built upon RefineNet~\cite{Lin2016RefineNet} with ResNet152 as backbone, unless specified otherwise. In the training stage, we set the momentum as $0.9$ and weight decay as $5\times 10^{-4}$. Following previous paper~\cite{Nekrasov_2019}, we train RGB and depth branches together with the same learning rate $10^{-3}$. The learning rate is divided by $10$ when the loss stops decreasing. The number of training epochs for NYUDv2 and SUNRGBD are 120 and 60 respectively. The size of the training batch is $8$, and the input image is resized to $640\times 480$.
We use random horizontal flipping, cropping, and image color augmentation. We report results based on three evaluation metrics, \emph{i.e.}, pixel accuracy, mean accuracy and mean Intersection over Union (IoU). We set $\alpha=1,\beta=0.1$ in (\ref{eqn:HPM}) according to cross validation.

\subsection{Correlation Analysis}\label{correlation}

As claimed in Section~\ref{sec:intro}, depth prediction error and segmentation error might be positively correlated. We provide a quantitative analysis of this hypothesis for both multi-task learning baseline (\emph{i.e.}, $L_s+L_d$) and our method based on Depth Prediction Error (DPE) (\emph{i.e.}, $L_s+L_d+L_h(\mathbf{R})$), in which ``$L_h(\mathbf{R})$" means we only use DPE map in $L_h$. Specifically, for test images in NYUDv2, we divide the logarithm of depth prediction error into $10$ bins and calculate the segmentation error rate of the pixels in each bin. The segmentation errors in all bins for two methods are plotted in Fig.~\ref{fig:cons}.

One observation is that the segmentation error rate grows higher as the depth prediction error increases, which proves that segmentation error and depth prediction error are positively correlated. Another observation is that our method beats multi-task learning in each bin, especially when the depth prediction error is large, which demonstrates the effectiveness of DPE.
We also compare our method ($L_s+L_d+L_h (\mathbf{R})$) with hard pixel mining baselines based on segmentation error in Section~\ref{sec:exp_hard_mining}, which shows that $L_h (\mathbf{R})$ performs more favorably probably because additional depth information provides extra valuable supervision.


\subsection{Comparison with the State-of-the-art}\label{sec:RDepth}

We divide all baselines into three groups: 1) \emph{S} group: RGB segmentation baselines without using depth information; 2) \emph{D} group: RGBD segmentation baselines using depth information for both training and test images; 3) \emph{P} group: depth privileged segmentation baselines using depth information only for training images. For a fair comparison, we also list the backbones of all methods. We report two versions of our method with backbone ResNet50 and ResNet152 in Table~\ref{stateofart}.

For the \emph{S} group, we compare with RefineNet~\cite{Lin2016RefineNet}, PSPNet~\cite{zhao2017pyramid}, and DeepLabv3~\cite{chen2017rethinking}.
As our backbone network RefineNet only releases MATLAB code, we implement it using python and obtain better results than those reported in\cite{Lin2016RefineNet}.

For the \emph{D} group, we compare with FCN~\cite{long2015fully}, LS-DeconvNet~\cite{cheng2017locality}, Wang \emph{et al.}~\cite{wang2018depth}, ACNet~\cite{hu2019acnet}, CFN~\cite{Di2017Cascaded}, and RDFNet~\cite{Lee2017RDFNet}. Among them, RDFNet is also built upon RefineNet and closely related to our method, so we additionally report our reproduced results.

For the \emph{P} group, depth information is not available in the testing stage. RDFNet is originally an RGBD segmentation method (D group), and we adapt it to use only RGB images during testing. In particular, we remove its depth input and retrain the remaining network for several epochs. We also try feeding both RGB images and pseudo depth maps predicted by our method to RDFNet during testing, which is marked with ``PD". LW-RefineNet~\cite{nekrasov2018lightweight}, TRL~\cite{zhang2018joint}, and PAP~\cite{zhang2019pattern} are multi-task learning method which can naturally use depth as privileged information. Among them, PAP~\cite{zhang2019pattern} used extra Surface Norm (SN) annotations as the third branch of multi-task learning during training, and thus achieves the best pixel-wise accuracy on SUNRGBD. For other methods from the \emph{D} group, some cannot be easily adapted to depth privileged method while some adapted ones are much worse than their original version, so we omit their results here.

Experimental results are summarized in Table~\ref{stateofart}. We observe that RefineNet~\cite{Lin2016RefineNet} achieves very competitive results within the \emph{S} group. We also observe that recent methods CFN~\cite{Di2017Cascaded}, ACNet~\cite{hu2019acnet}, and RDFNet~\cite{Lee2017RDFNet} from \emph{D} group are generally better than the methods in \emph{S} group, which proves that depth information is beneficial for semantic segmentation. Our method achieves significant improvements over our backbone network RefineNet, \emph{i.e.}, 4.17\% and 4.92\% IoU improvement on SUNRGBD and NYUDv2 datasets respectively. Besides, our method generally beats the baselines in the \emph{P} group. Although PAP~\cite{zhang2019pattern}(+SN) uses extra Surface Norm (SN) annotations during training, our method is still better than \cite{zhang2019pattern} except pixel accuracy on SUNRGBD. Moreover, our method also generally outperforms the baselines in the \emph{D} group, which use more information than our method during testing.

%
%
%
%

\subsection{Ablation Studies}

To validate the effectiveness of each component in our method, by taking NYUDv2 as an example, we report the results of our special cases in Table~\ref{abalation}. The first four rows are baselines without using our LWM. Specifically, $L_{s}$ is RefineNet152 in Table~\ref{stateofart} which only uses segmentation loss. $L_{s}+L_{d}$ is a multi-task learning baseline. We futher try using multi-scale (MS) training (calculate losses on multi-scale side outputs) or curriculum learning (CL) based on $L_{s}+L_d$. In the next three rows, we report the results of $L_{s}+L_{d}+L_h$ using only DPE ($\mathbf{R}$), only DSE ($\mathbf{Z}$), or both in $L_h$. Finally, we report the results of $L_{HPM}$ and $L_T$ by adding MS and CL to $L_{s}+L_{d}+L_h$ step by step.


\begin{table}[]
\centering
\caption{IoUs (\%) of our \textbf{special cases} on NYUDv2. MS (\emph{resp.}, CL) is short for Multi-Scale training (\emph{resp.}, Curriculum Learning).}\label{abalation}
\begin{tabular}{cc|cc|cc|c}
\toprule[1.5pt]
$L_{s}$   & $L_{d}$  &$L_h(\mathbf{R})$ & $L_h(\mathbf{Z})$ &MS & CL  & \textbf{IoU}  \\ \hline \hline
\checkmark             &              &      &                                   &                      &          & 46.59 \\
\checkmark             &\checkmark   &      &                                   &                      &          &47.25 \\
\checkmark             &  \checkmark         &         &                                   &                   \checkmark   && 47.51 \\
\checkmark             &  \checkmark            &      &                                   &                      &\checkmark& 47.84 \\ \hline
\checkmark             &\checkmark &\checkmark& &                                   &          &49.15\\
\checkmark             & \checkmark             &      &   \checkmark                                &&          &49.63\\
\checkmark             &\checkmark & \checkmark       &\checkmark &                                 &          &50.31\\
\checkmark             &\checkmark &\checkmark&       \checkmark         &\checkmark                      &    &  50.93     \\ \hline
\checkmark             &\checkmark  &\checkmark&    \checkmark         &\checkmark&\checkmark&\textbf{51.51}\\ \bottomrule[1.5pt]
\end{tabular}
\label{tab}
\end{table}

\begin{table}[]
\caption{IoUs (\%) of our method ($L_{HPM}$) on NYUDv2 when using different fusion functions or hyper-parameters.}
\centering
\subcaption{\textbf{Different fusion functions}.}
\setlength{\tabcolsep}{16pt}
\begin{tabular}{c||c|c|c}
    \hline
    $F(\cdot)$      & $max$ & $*$     &$\bm{+}$ \\ \hline \hline
    \textbf{IoU}&49.69&48.55&\textbf{50.93}\\ \hline
\end{tabular}
\vspace{0.2cm}

\label{hyperf}
\subcaption{\textbf{Different numbers of cells $m$}.}
\setlength{\tabcolsep}{10pt}
\begin{tabular}{c||c|c|c|c}
\hline
$m$      & $1\times 1$ & $w\times h$ &  $16\times 16$ &\bm{$8\times 8$} \\ \hline \hline
\textbf{IoU}&49.63&50.04 &50.42&\textbf{50.93}\\ \hline
\end{tabular}
\vspace{0.2cm}

\label{hyperm}
\subcaption{\textbf{Different bin sizes $d$}.}
\setlength{\tabcolsep}{11pt}
\begin{tabular}{c||c|c|c|c}
\hline
$d$      & 1 & \textbf{10} &  50 &100 \\ \hline \hline
\textbf{IoU}&50.04 &\textbf{50.93} &50.68 &50.21\\ \hline
\end{tabular}
\label{hyperd}
\label{hyperd}
\end{table}

\setlength{\tabcolsep}{10pt}
\begin{table*}[]
\centering
\caption{IoUs (\%) of different \textbf{hard pixel mining} methods based on $L_{s}+L_d$ on NYUDv2.}
\label{tab:hard_pixel_mining}
\begin{tabular}{c|c|c|c|c|c|c}
\hline
 & $+$OHEM~\cite{Shrivastava2016Training}  & $+$LMP~\cite{bulo2017loss} &  $+$FL~\cite{Lin2017Focal} & $+L_h(\mathbf{R})$ & $+L_h(\mathbf{Z})$  &$+L_h$ \\ \hline \hline
\textbf{IoU}&47.25 &47.46&48.32& 49.15& 49.63&\textbf{50.31}\\ \hline
\end{tabular}
\end{table*}


From Table~\ref{abalation}, we observe that the improvements of adding multi-scale (MS) and curriculum learning (CL) to multi-task baseline $L_s+L_d$ is minor. Moreover, the largest performance gain is brought by our LWM ($L_{s}+L_{d}+L_h$ \emph{v.s.} $L_{s}+L_{d}$), about 3.06\%. Our proposed method is more effective compared with multi-task learning ($L_{s}+L_{d}$ \emph{v.s.} $L_{s}$), which only lead to 0.66\% IoU improvements. It can also be seen that DPE ($\mathbf{R}$) and DSE ($\mathbf{Z}$) are both useful, while $\mathbf{Z}$ is more effective than $\mathbf{R}$. Besides, using both DPE and DSE is better than using DPE or DSE alone, which reflects that our proposed two measurements are complementary with each other and can jointly contribute to the final performance. Finally, multi-scale $L_{hk}$ and curriculum learning can also help achieve better results, which demonstrates the effectiveness of applying LWM to multi-scale side outputs and our designed training strategy.

\subsection{Comparison with hard pixel mining methods} \label{sec:exp_hard_mining}

By taking NYUDv2 as an example, we further list the results of applying OHEM~\cite{Shrivastava2016Training}, focal loss (FL)~\cite{Lin2017Focal}, and LMP~\cite{bulo2017loss} to multi-task learning baseline $L_{s}+L_{d}$ in Table~\ref{tab:hard_pixel_mining} for comparison because they all use only RGB images or segmentation losses to mine the hard training pixels. Specifically, LMP assigns larger weights on pixels with larger losses by proposed max-pooling function, while OHEM only calculates the losses of pixels which are larger than a threshold and ignore the remaining ones. Recently, focal loss down-weights the loss assigned to well-segmented pixels. For our method, we copy three results from Table~\ref{tab:hard_pixel_mining}: $L_s+L_d+L_h(\mathbf{R})$, $L_s+L_d+L_h(\mathbf{Z})$, and $L_s+L_d+L_h$.

From Table~\ref{tab:hard_pixel_mining}, it can be seen that our method with single-scale LWM is more effective than merely mining hard pixels based on RGB information ($L_{s}+L_{d}+L_h$ \emph{v.s.} $L_{s}+L_{d}+$OHEM/FL/LMP), which indicates the advantage of fusing RGB and depth to mine hard pixels. Moreover, when only using $\mathbf{R}$ or $\mathbf{Z}$ for $L_h$, our method is still better than other hard pixel mining baselines.

\subsection{Hyper-parameters Analyses}\label{sec:hyper}
\begin{figure}[]
\centering
\includegraphics[width=\linewidth]{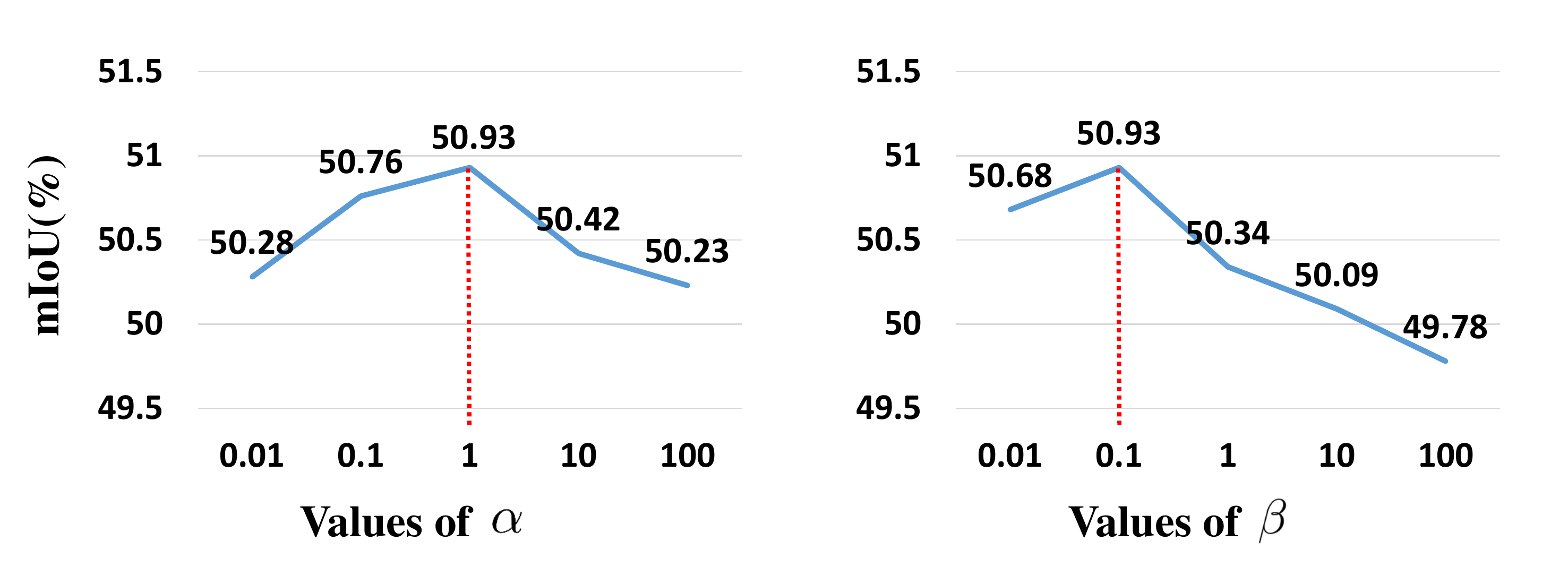}
\caption{IoUs (\%) of our method ($L_{HPM}$) when varying \textbf{hyper-parameters} $\alpha,\beta$ on NYUDv2. The dashed lines denote the default values used in our paper. }\label{alpha}
\end{figure}

In this section, we investigate the impact of different hyper-parameters in our method by taking NYUDv2 dataset as an example. We conduct experiments using our method ($L_{HPM}$) without curriculum learning, considering that $L_{HPM}$ is our main contribution.
Recall that we have two hyper-parameters $\alpha$ and $\beta$ in (\ref{eqn:T}).
We plot the performance variance when changing the hyper-parameters $\alpha,\beta$ in Fig.~\ref{alpha}. When varying $\alpha$ (\emph{resp.}, $\beta$) within the range [0.01,100], and the obtained IoU (\%) of our method ($L_{HPM}$) is in the range of [50.23, 50.93] (\emph{resp.}, [49.78, 50.93]) as shown in Fig.~\ref{alpha}, which demonstrates the effectiveness of our method when setting $\alpha$ and $\beta$ in a reasonable range.

\begin{table}
\centering
\caption{IoUs (\%) of different methods using different \textbf{base networks} on NYUDv2.}
\begin{tabular}{c|ccc}
\toprule[1.5pt]
Base networks   & $L_s+L_d$ &$L_s+L_d+$FL&$L_s+L_d+L_{h}$    \\ \hline \hline
Deeplabv3 & 47.23& 48.56&50.02  \\
PSPNet  & 47.19 & 48.20&49.56 \\
RefineNet& 47.25&48.32 &\textbf{50.31}  \\ \bottomrule[1.5pt]
\end{tabular}
\label{backbone}
\end{table}

\begin{figure*}
\centering
\includegraphics[width=\linewidth]{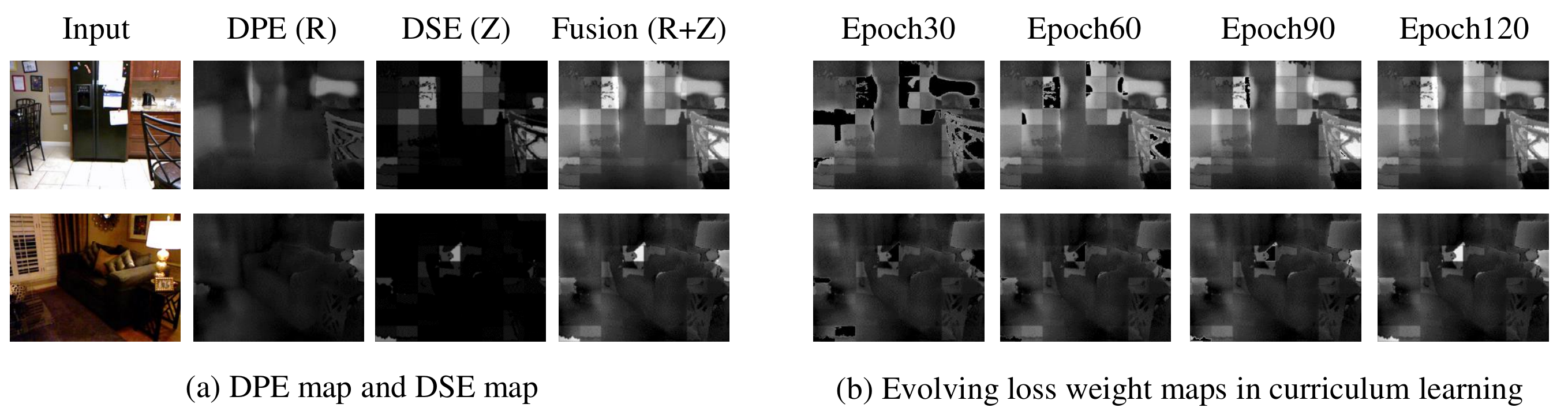}
\caption{\textbf{Visualizations of our loss weight map}. The subfigure (a) shows the visualization of Depth Prediction Error (DPE) map ($\mathbf{R}$), Depth-aware Segmentation Error (DSE) map ($\mathbf{Z}$), and the fused map $\mathbf{M}$ using pixel-wise summation. The subfigure (b) shows the loss weight map evolving with the training epoch in curriculum learning. Note that the brighter pixels correspond to larger weights, which indicate harder pixels.}
\label{vis}
\end{figure*}

In Section~\ref{sec:fusion_strategy}, we mentioned that different fusion functions $F(\cdot)$ can be used to fuse DPE map and DSE map. Here, we explore three simple strategies: pixel-wise max, pixel-wise product, and pixel-wise summation. The results are summarized in Table~\ref{hyperf}, which shows that pixel-wise summation achieves the best result.

In Section~\ref{sec:DSE}, we have bin size $d$ when grouping depth values into bins and the number of cells $m=\bar{w}\times\bar{h}$ when dividing the whole image into cells. The results using different $d$ are reported in Table~\ref{hyperd}, which indicates that our method is insensitive to the bin size $d$.
Finally, we report the results using different $m$ in Table~\ref{hyperm}. It can be seen that the results with $m=8\times 8$ are better than those with $m=1\times1$ or $m=w\times h$, which are two extreme cases as discussed in Section~\ref{sec:DSE}. The results verify that it would be more beneficial to assign higher weights on hard local regions instead of hard global regions ($8\times 8$ \emph{v.s.} $1\times 1$) or only misclassified pixels ($8\times 8$ \emph{v.s.} $w\times h$).

\subsection{Base Network Analyses}
Note that our method is built upon RefineNet as the base network for segmentation.
However, our hard pixel mining method only requires additional depth prediction branch and thus can be adapted to other base networks with minor modifications. In this section, we attempt to build our method on other RGB segmentation networks like Deeplabv3 and PSPNet by adding a $1\times 1$ conv layer after their last pooling layer to predict depth. By using three different base networks with backbone ResNet152, we report the results of multi-task learning baseline $L_s+L_d$, our method ($L_s+L_d+L_h$) with single-scale LWM (see Table~\ref{tab}), and a competitive hard pixel mining baseline $L_s+L_d+FL$ (see Table~\ref{tab:hard_pixel_mining}) on NYUDv2.

The performances on NYUDv2 are shown in Table~\ref{backbone}, from which we can observe the effectiveness of our method ($L_s+L_d+L_h$) compared with the multi-task baseline and the hard pixel mining baseline when using different base networks. These results indicate the good generalization ability of our method.

\begin{figure*}[!htbp]
\centering
\includegraphics[width=0.9\linewidth]{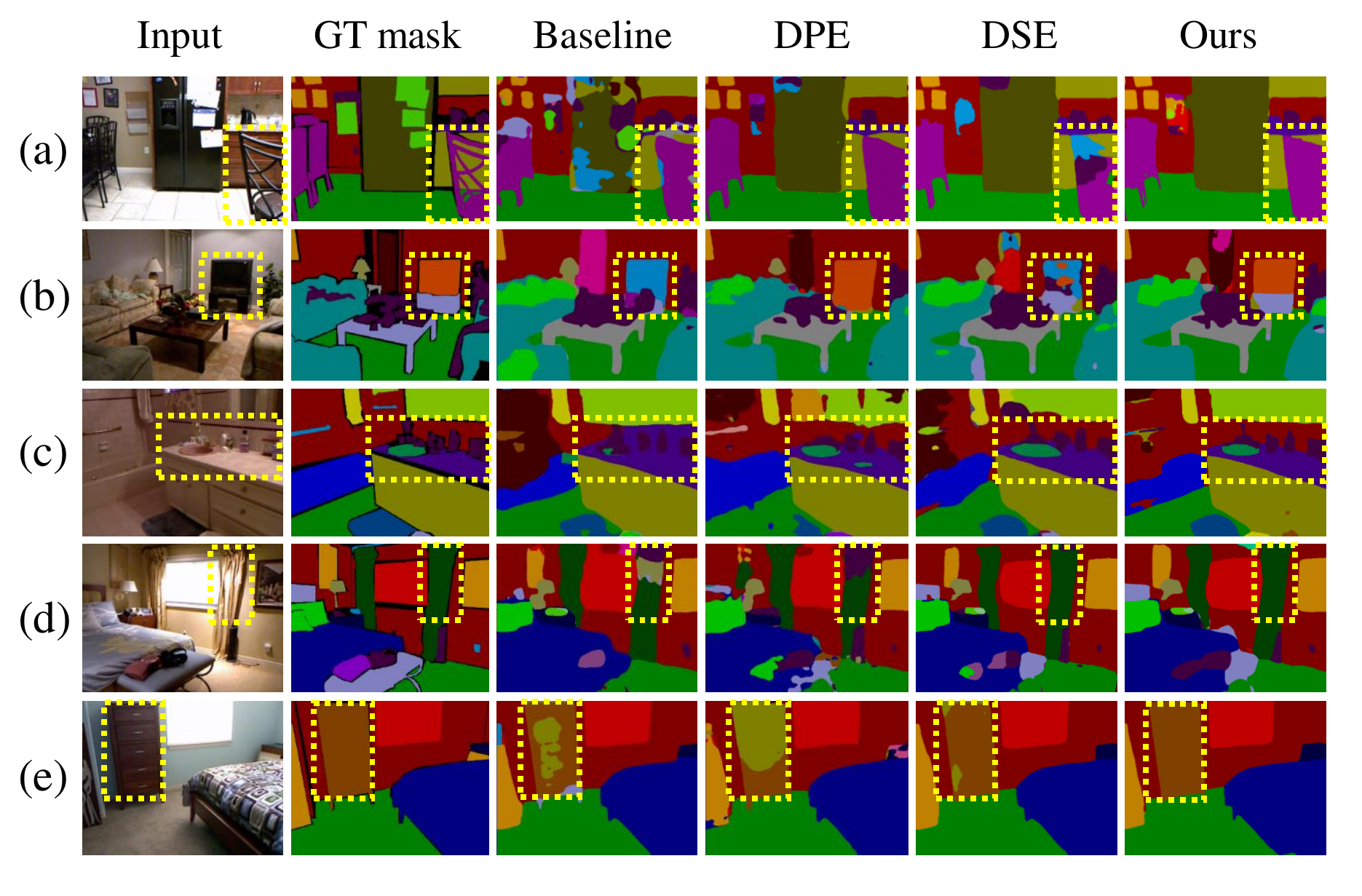}
\caption{\textbf{Qualitative results} of our special cases on the NYUDv2 test set. From left to right: the input images, the ground-truth segmentation masks, the segmentation results of our base network RefineNet ($L_s$), the results after using DPE ($L_s+L_d+L_h(\mathbf{R})$), the results after using DSE ($L_s+L_d+L_h(\mathbf{Z})$), and the results of our full method. The bounding boxes highlight the improvements of our method. Best viewed in color.}
\label{good}
\end{figure*}

\begin{figure*}[!htbp]
\centering
\includegraphics[width=0.9\linewidth]{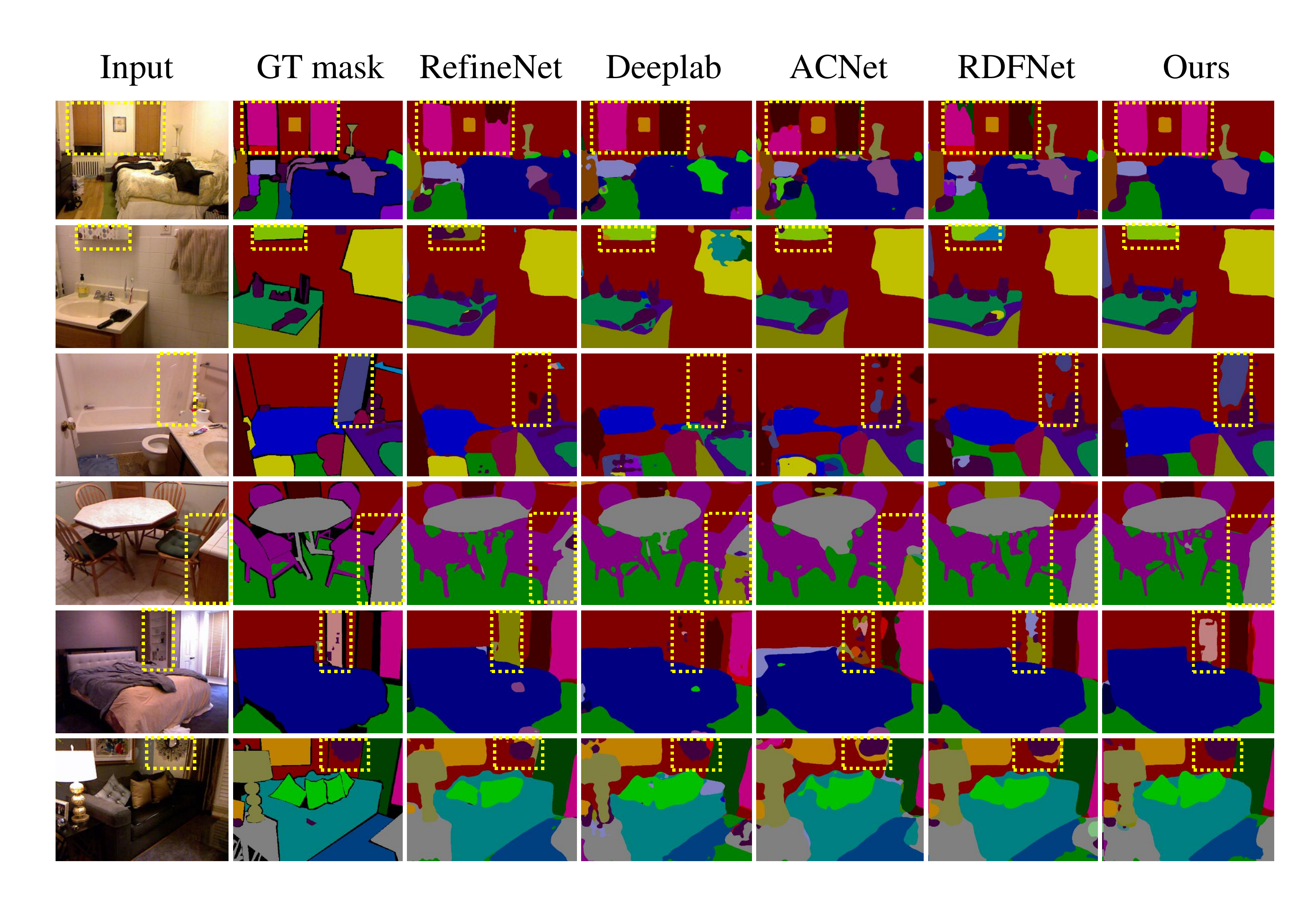}
\caption{\textbf{Qualitative results} of state-of-the-art methods and our method on the NYUDv2 test set. From left to right: the input images, the ground-truth segmentation masks, the segmentation results of RefineNet, Deeplabv3, ACNet, RDFNet, and our method. The bounding boxes highlight the improvements of our method. Best viewed in color.}
\label{new}
\end{figure*}

\subsection{Visualization of Loss Weight Map}

By taking two images from NYUDv2 dataset as examples, we visualize the loss weight maps obtained by our method in Fig.~\ref{vis}, in which the brighter pixels correspond to larger weights and indicate harder pixels. As shown in Fig.~\ref{vis} (a), we list the input images, Depth Prediction Error (DPE) maps ($\mathbf{R}$), Depth-aware Segmentation Error (DSE) maps ($\mathbf{Z}$), and the fused loss weight maps $\mathbf{M}$ using pixel-wise summation. It can be seen that DPE map and DSE map have certain overlap because depth prediction error is positively correlated with segmentation error as shown in Section~\ref{correlation}. However, DPE focus on the hard pixels with huge depth gap while DSE could identify some hard regions not specified by DPE, which proves that DPE and DSE are complementary with each other and jointly lead to the performance gain.

Moreover, to visualize the procedure of curriculum learning, we exhibit the loss weight maps $\delta(M_i<\eta_e)M_i$ (see (\ref{eqn:cl_m})) in different epochs (\emph{e.g.}, 30, 60, 90, 120) from a total of $120$ training epochs in Fig.~\ref{vis} (b). As the training epoch increases, more bright pixels appear in the loss weight map, which means that more hard pixels participate in the training procedure and contribute to learning a more robust model.

\subsection{Qualitative Analyses}

In this section, we provide some visualization results of the baselines and our method on the NYUDv2 test set. To fully corroborate the effectiveness of our method, we visualize both the comparison with state-of-the-art methods in Fig.~\ref{new} and the comparison with our special cases in Fig.~\ref{good}.

In Fig.~\ref{good}, we visualize the segmentation results of our base network RefineNet ($L_s$), the results after using DPE ($L_s+L_d+L_h(\mathbf{R})$), the results after using DSE ($L_s+L_d+L_h(\mathbf{Z})$), and the results of our full method. In the ``DPE" column, it can be seen that our method is adept at segmenting some objects with sharp depth boundary (a,b). For example, our method can successfully segment the upper part of the chair back in (a), the table under the TV set in (b), which is attributed to our DPE map. In the ``DSE" column, our method also demonstrates excellent ability in classifying confusing objects with similar visual appearances in the same depth bin (c,d,e), which are misclassified as incorrect categories by baseline RefineNet. For example, our method can successfully segment the sink in (c), the curtain in (d), and the cabinet in (e), which is credited to our DSE map. In the ``Ours" column, we can also observe that by combining DPE and DSE to mine the hard pixels, our model achieves significantly better results.

In Fig.~\ref{new}, we visualize the segmentation results of different state-of-the-art methods. The results in Fig.~\ref{new} further show that our method significantly outperforms other RGB and even RGBD segmentation methods, especially on depth-related regions such as mirror, armoire, and desk.

\section{Conclusions}
In this work, we have proposed to mine hard pixels by fusing RGB information and depth information in depth privileged semantic segmentation task. In particular, we have designed a novel Loss Weight Module (LWM), which fuses two measurements of hard pixels (depth prediction error and depth-aware segmentation error) and outputs a loss weight map for weighting pixel-wise training losses.
Moreover, we have applied LWM to multi-scale side outputs and explored curriculum learning strategy based on LWM. Comprehensive quantitative and qualitative results on two benchmark datasets have demonstrated the superiority of our method.

\ifCLASSOPTIONcaptionsoff
  \newpage
\fi


\bibliographystyle{IEEEtran}
\bibliography{egbib}
\end{document}